\documentclass[a4paper,fleqn]{cas-dc}

\usepackage[numbers,sort&compress]{natbib}
\usepackage{amsmath,amssymb,amsfonts}
\usepackage{booktabs}
\usepackage{graphicx}
\usepackage{subcaption}
\usepackage{algorithm}
\usepackage{algpseudocode}
\usepackage{caption}
\usepackage{needspace}
\usepackage{float}
\usepackage{placeins}
\usepackage{algorithm}
\usepackage{algpseudocode}
\usepackage{caption}
\usepackage{stfloats}
\usepackage{cuted}
\usepackage{subcaption}
\usepackage{cleveref}
\newtheorem{proposition}{Proposition}
\newif\ifshowedits
\showeditstrue

\ifshowedits
  
\else
  
\fi

\definecolor{finalrevisioncolor}{HTML}{8054A2}
\newif\ifshowgreenedits
\showgreeneditstrue

\ifshowgreenedits
  \newcommand{\grev}[1]{\begingroup\color{black}#1\endgroup}
\else
  \newcommand{\grev}[1]{#1}
\fi
\newif\ifshowpurpleedits
\showpurpleeditstrue

\ifshowpurpleedits
  
\else
  
\fi

\begin{document}
\let\WriteBookmarks\relax
\def\floatpagepagefraction{1}
\def\textpagefraction{.001}
\shorttitle{Geometry-Based Feature Construction for Image Classification}
\shortauthors{Mauree and Arya}

\title[mode=title]{Beyond Landmark Extraction: A Framework for Robust Geometric Feature Evaluation in Structured Image Classification}

\author[1]{Saravana Mauree}
\cormark[1]
\ead{sxm2176@case.edu}
\credit{Conceptualization, Methodology, Software, Formal analysis,
Investigation, Data curation, Visualization, Writing -- original draft}

\author[1]{Sakshi Arya}
\ead{sxa1351@case.edu}
\credit{Conceptualization, Methodology, Supervision, Validation,
Writing -- review \& editing}

\affiliation[1]{
organization={Case Western Reserve University},
addressline={Department of Mathematics, Applied Mathematics, and Statistics},
city={Cleveland},
state={OH},
country={USA}
}

\cortext[1]{Corresponding author}
\RenewDocumentCommand{\printorcid}{}{}
\begin{abstract}
Much of the literature on structured image recognition has disproportionately focused on the comparison of classification algorithms. Rather than investigating which classifier performs best, this paper instead asks: what should a classifier know before it ever makes a prediction? In structured vision problems such as gesture recognition, facial expression categorization, and medical image analysis, discriminative information lies less in individual pixels and more in spatial relationships between semantic parts. Raw pixel spaces are high-dimensional, sensitive to nuisance variation, and often obfuscate the geometric structures that make visual tasks interpretable. Landmark extraction provides one form of dimension reduction, but it does not by itself determine the information preserved. This paper studies the post-landmark feature map as the central object of analysis and proposes a systematic framework for constructing and interpreting landmark-derived representations as an, informed, feature-based ``dimension reduction'' step. Using static hand gesture recognition as a case study, we evaluate coordinate, distance, angle, and hybrid representations through \grev{image-corruption robustness and component-ablation experiments under participant-disjoint evaluation}. The results show that visually variable data exposes substantial gaps between raw coordinate features and their geometrically invariant counterparts, while hybrid representations achieve the strongest overall performance by combining complementary geometric components. \grev{We further establish that distinct representations can preserve equivalent geometric information despite differing in dimensionality and predictive performance.}These findings frame feature construction as a fundamental modeling decision and ultimately suggest that the question of what representations should a classifier learn from is one worth asking. The code used for feature construction and evaluation is available at \href{https://github.com/ShivMaureeCWRU/Feature_based_dimension_reduction}{GitHub Repository}
\begin{center}
\vspace{-1em}
\small\textit{This manuscript is under consideration at Pattern Recognition Letters.}
\vspace{-2em}
\end{center}
\end{abstract}

\maketitle

\section{Introduction}
\label{sec:introduction}
Image classification is often formulated in its most direct form: building a map from raw pixel data to a discrete set of class labels. Images are represented through high-dimensional arrays of unprocessed pixel intensities, and the classifier is tasked with inferring relevant geometric structures from these raw observations. Pixel-based methods, notably convolutional neural networks, have driven momentous advances 
in computer vision \citep{Krizhevsky2012,LeCun2015}. However, these seemingly irreproachable methods come with one caveat: raw pixel representations bear well-known limitations. They are high-dimensional, sensitive to nuisance variation including scale, illumination, and capture point, and are often difficult to interpret \citep{Lowe2004,Guidotti2018}. For many structured vision tasks, the discriminative information intrinsic to classification is not uniformly distributed across the image. Instead, it is often concentrated within spatial configurations of visually perceptible features; a concept perfectly illustrated through static hand gesture recognition. Here the class label is determined less by the layout of the background or the lighting of the scene than by the relative positioning of the fingers, joints, and palm. Similar considerations arise with facial expression recognition where classes are defined through relationships amongst facial landmarks, and in pose-based activity recognition, where relevant class information is largely contained within body configuration. Landmark detectors provide a practical bridge between raw images and structured geometric representations. MediaPipe Hands estimates a $21$-landmark hand skeleton from a single RGB camera input, producing a remarkably compact representation of hand shape and pose \citep{Zhang2020MediaPipeHands}. Moreover, these landmark-based pipelines:
\begin{center}
\vspace{-0.5em}
\begin{minipage}{0.92\linewidth}
\centering
\includegraphics[width=\linewidth]{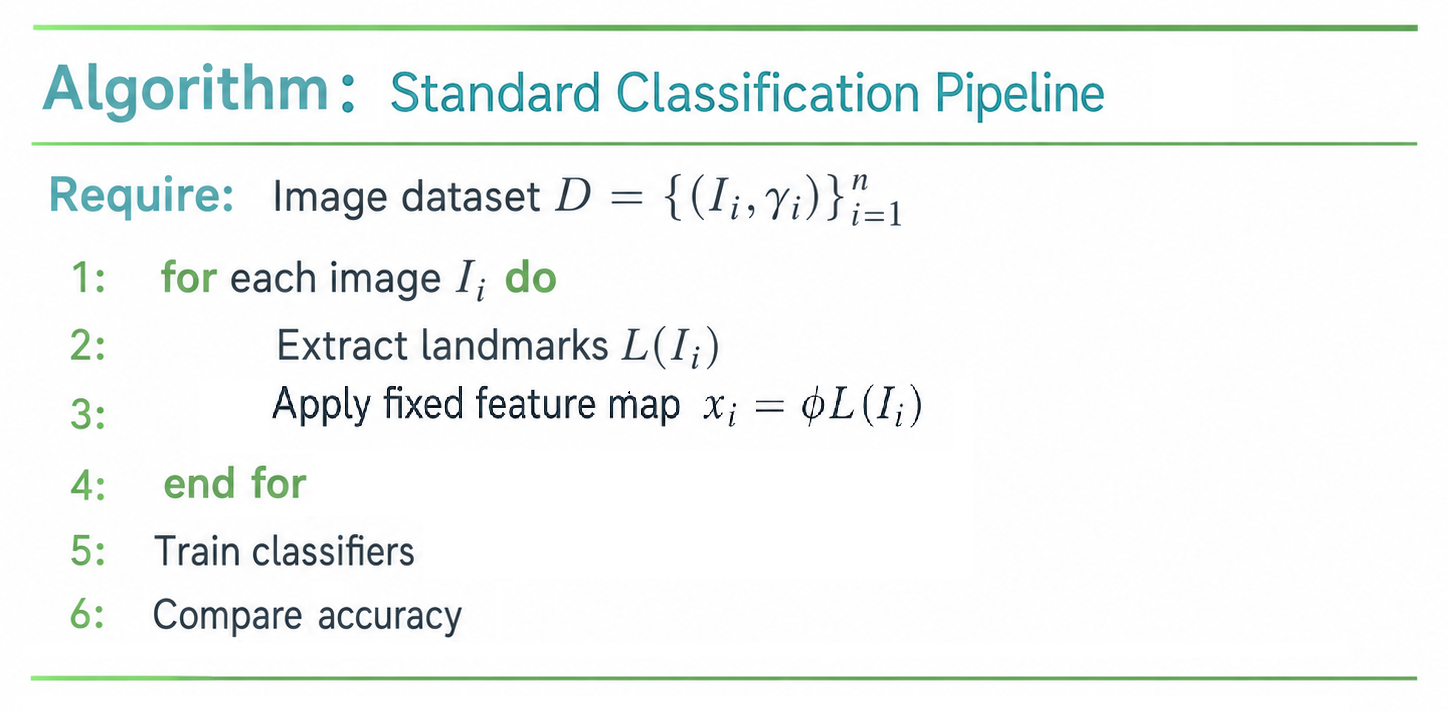}
\captionsetup{type=algorithm, labelformat=empty, textformat=empty}
\captionof{algorithm}{Framework Algorithm}
\label{fig:algorithm_standard}
\end{minipage}
\vspace{-4em}
\end{center}
 have shown varying levels of success in real time hand gesture recognition, through neural networks and other classifiers \citep{Sung2021Gesture,Verma2024MediapipeCNN}.
Whilst this conventional pipeline illustrated in Algorithm~\ref{fig:algorithm_standard} has been effective, it overlooks a central modeling choice: what form of information should the classifier be exposed to? Should it see absolute landmark positions, distances between joints, normalized hand shape, local finger angles, or some combination of these structures? These choices are distinct and not necessarily interchangeable. Each feature map encodes a different modeling assumption that dictates the geometry preserved or the nuisance \grev{variation} removed. We refer to this post-landmark construction as problem-informed dimension reduction in the broader sense that the image is first condensed to semantic structure and then transformed to retain task-relevant geometry whilst suppressing nuisance variability. In this sense, the reduction is defined through the effective variation carried forward from the original image representation, rather than strict feature-count reduction for every candidate map. The feature map therefore determines whether the classifier is guided toward gesture-discriminative geometry or left to model irrelevant variation. \grev{Geometric normalization and feature construction have been extensively investigated in computer vision, with invariant descriptors and geometry-based approaches abetting hand gesture recognition \citep{Lowe2004,Priyal2010}. Whilst these studies underscore the importance of representation design, they further motivate a systematic comparison of alternative features derived from a common set of landmarks.} The framework developed in this paper \grev{takes this comparison forward by} treating the representation map itself as the primary modeling object. \grev{Using these relatively conventional geometric descriptors as a basis, we systematically examine three complementary aspects of representation design: through classification comparisons, we evaluate the influence of representation choice on predictive performance; through perturbation experiments, we investigate variations in end-to-end robustness under image degradation; and through component ablations, we assess the incremental predictive value of the individual features}.

This perspective \grev{of evaluating problem-specific geometric feature representations} is not unique to gesture recognition. In medical image analysis, diagnostically relevant data is often encoded in the shape, orientation, and relative positioning of anatomical structures rather than in the unprocessed pixel space or voxel fields \citep{Aerts2014, Gillies2016}. Once such structures are extracted through segmentations and contours, the modeling decision shifts from whether the structure is extractable to how it should be represented \citep{Lambin2017}. Distances, angles, coordinates, and hybrid geometric summaries may correspond to different anatomical assumptions, and their interpretability is crucial in a setting where the model's "reasoning" must be explained \citep{Tjoa2020}. \grev{Static gesture recognition therefore serves as a controlled setting for examining the influence of representation choice on classification performance. Through a common extraction pipeline, fixed classifiers, and the intentional exclusion of temporal dependencies, we isolate the influence of spatial geometric inclusion on predictive performance, foregoing the usual emphasis on detector architecture and classifier optimization.} \grev{Amongst the standard geometric primitive is a} hybrid representation that combines \grev{axis-normalized} coordinates, normalized pairwise distances, and angle features to encode global hand layout, relative landmark spacing, and local finger geometry. \grev{We also consider an angle-free simple hybrid representation and a centered, diameter-normalized coordinate representation. Proposition~\ref{prop:geometric_equivalence}  establishes that these two representations determine one another on noncollinear configurations, distinguishing the geometry preserved from the encoding supplied to a classifier.}

The contributions of this work are fourfold. First, we formalize post-landmark feature construction as an interpretable, problem-informed dimension reduction step in which task geometry guides what information is preserved and what nuisance variation is suppressed. Second, we provide a systematic framework for evaluating \grev{and selecting} landmark-derived representations\grev{, together with a characterization of the geometric information preserved by the simple hybrid} \grev{representation}. Third, through perturbation and ablation analyses, we \grev{assess image-corruption robustness and the incremental predictive contributions of individual feature components}. Fourth, we show how these analyses can guide the principled construction of hybrid representations\grev{, whilst distinguishing information preservation from predictive utility under the chosen classifiers}.

\vspace{-1.2em}
\section{Problem Formulation}
\label{sec:Problem Formulation}
Let
\begin{center}
\vspace{-2.7em}
 $\mathcal{D}=\{(I_i,\gamma_i)\}_{i=1}^n$,   
 \vspace{-0.7em}
\end{center}
denote a labeled image dataset, where $I_i$ is the $i$-th image and $\gamma_i\in\{1,\ldots,K\}$ is its class label. A standard pixel-based classifier attempts to learn
\vspace{-1em}
\begin{center}
   \hspace*{1cm}
   $f:\mathbb{R}^{H \times W \times C}\rightarrow \{1,\ldots,K\},$
\vspace{-0.7em}
\end{center}
where $H$, $W$, and $C$ denote image height, width, and number of color channels. Here, classification is separated into three stages:
\begin{center}
\vspace{-1.7em}
\hspace*{1cm}$
I_i\longrightarrow L(I_i)
\longrightarrow Q_i
\longrightarrow \phi_r(Q_i)
\longrightarrow g(\phi_r(Q_i)),
$
\vspace{-0.5em}
\end{center}
where $L:\mathbb{R}^{H\times W\times C}\rightarrow \mathbb{R}^{21\times d}$where $L:\mathbb{R}^{H\times W\times C}
\rightarrow\mathbb{R}^{21\times d}$ extracts
the hand landmarks, with $d$ retained coordinates, $Q_i\in\mathbb{R}^{21\times 2}$
denotes their aspect-ratio-corrected planar configuration,
and $\phi_r:\mathbb{R}^{21\times 2}
\rightarrow\mathbb{R}^{q_r}$ constructs the
$r$-th feature representation. The transformed dataset associated with $\phi_r$ is
$\mathcal{D}_{\phi_r}
=\{(\phi_r(Q_i),\gamma_i)\}_{i=1}^n$
This formulation shifts the modeling emphasis
from classifier comparison under a fixed representation to the geometric information preserved by $\phi_r$ and its predictive utility.

\vspace{-0.5em}
\section{Landmark feature representations}
\label{sec:landmark feature representations}
\begin{strip}
\vspace{-1.68em}
\begin{center}
\includegraphics[width=0.95\textwidth]{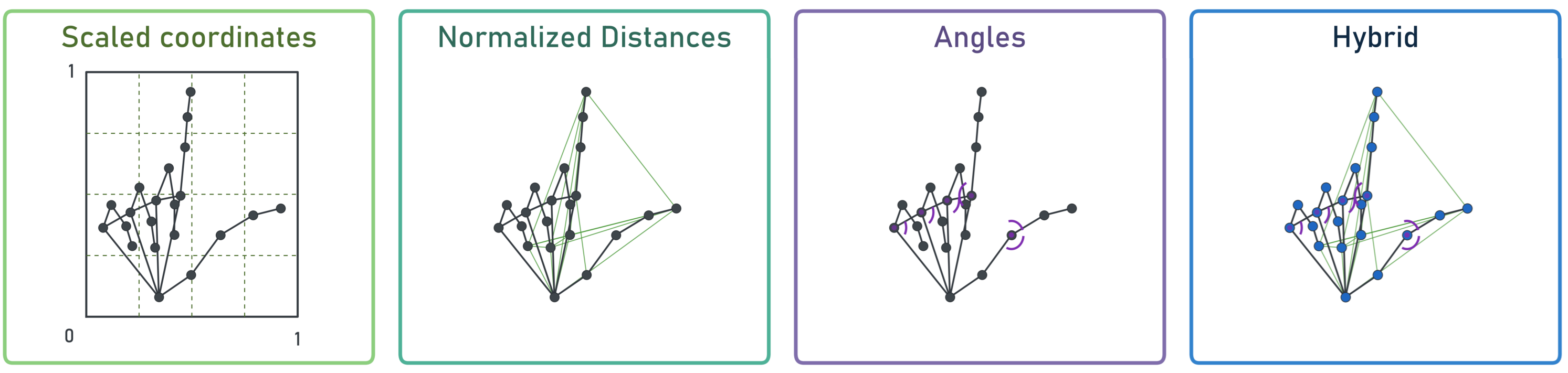}
\vspace{-1em}
\captionsetup{type=figure}
\captionof{figure}{Feature Visualisation-Summary And Justification of feature families considered}
\label{fig:Feature visualisation}
\vspace{-4em}
\end{center}
\end{strip}

\grev{Formally, let $L(I_i)=\{(x_j,y_j,z_j)\}_{j=0}^{20}$ denote the MediaPipe output for image $I_i$. Retaining the image-plane coordinates, we correct for aspect ratio to obtain a common Euclidean basis configuration $Q = \{q_j\}_{j=0}^{20}$, where
\begin{center}
\vspace{-0.8em}
\refstepcounter{equation}
\resizebox{0.6\linewidth}{!}{$
q_j=\frac{(W_i x_j,H_i y_j)}{\max(W_i,H_i)},
\qquad j=0,\ldots,20
\qquad (\theequation)
$}
\vspace{-0.5em}
\end{center}
and $W_i,H_i$ denote the image dimensions. From $Q$, we construct alternative feature maps $\phi_r(Q)$, illustrated in Figure~\ref{fig:Feature visualisation}. We define the coordinate minima $b_i$, axis-range matrix $\Lambda_i$, and landmark-set diameter $D_i$ as
\begin{center}
\vspace{-0.6em}
\refstepcounter{equation}
\resizebox{0.9\linewidth}{!}{$
b_i =
\begin{pmatrix}
\min_j q_{jx}\\
\min_j q_{jy}
\end{pmatrix},
\quad
\Lambda_i=
\begin{pmatrix}
R_x & 0\\
0 & R_y
\end{pmatrix},
\quad
D_i=\max_{0\leq a<b\leq20}\|q_a-q_b\|_2
\qquad (\theequation)
$}
\vspace{-0.5em}
\end{center}
where $R_k=\max_j q_{jk}-\min_j q_{jk}$ for $k\in\{x,y\}$. The raw-coordinate representation $\phi_q$ retains $Q$, while translated ($\phi_T$), axis-normalized ($\phi_A$), and centered isotropically normalized ($\phi_Z$) coordinates are constructed from
\begin{center}
\vspace{-0.8em}
\refstepcounter{equation}
\resizebox{0.75\linewidth}{!}{$
T_j= q_j-b_i, \quad A_j= \Lambda_i^{-1}(q_j-b_i),\quad Z_j= \frac{q_j-\bar q}{D_i},
\qquad (\theequation)
$}
\vspace{-0.8em}
\end{center}
where $\bar q=\frac{1}{21}\sum_{j=0}^{20}q_j$. Translation removes absolute position, axis normalization independently rescales both coordinate axes, and isotropic normalization preserves aspect ratio and orientation relative to the image frame. Pairwise distances and their normalized counterparts are defined as
\begin{center}
\vspace{-0.8em}
\refstepcounter{equation}
\resizebox{0.7\linewidth}{!}{$
d_{ab}=\|q_a-q_b\|_2,
\qquad
\delta_{ab}=\frac{d_{ab}}{D_i},
\quad 0\leq a<b\leq20
\qquad (\theequation)
$}
\vspace{-0.9em}
\end{center}
The representations $\phi_d$ and $\phi_\delta$ contain all pairs in lexicographic order, with normalization removing dependence on uniform scale. For angular features, we define
\begin{center}
\vspace{-0.8em}
\refstepcounter{equation}
\resizebox{0.6\linewidth}{!}{$
\theta_{abc}
=
\cos^{-1}\left(
\frac{(q_a-q_b)^\top(q_c-q_b)}
{\|q_a-q_b\|_2\|q_c-q_b\|_2}
\right)
\qquad (\theequation)
$}
\label{eq:angle}
\vspace{-0.5em}
\end{center}
The $\phi_\theta$ contains $15$ unsigned angles in radians, using the ordered triplets $(0,j,j+1)$, $(j,j+1,j+2)$, and $(j+1,j+2,j+3)$ for each finger start $j\in\{1,5,9,13,17\}$, capturing local finger articulation.The simple hybrid $\phi_s$ and full hybrid $\phi_f$ combine the geometric features through concatenation:$\phi_s(Q)=[\phi_A(Q),\phi_\delta(Q)]$ and $\phi_f(Q)=[\phi_A(Q),\phi_\delta(Q),\phi_\theta(Q)]$
Two additional block-removal representations are constructed for ablation only: $\phi_{f\setminus A}=(\phi_\delta,\phi_\theta)$, $\phi_{f\setminus\delta}=(\phi_A,\phi_\theta)$. Their invariances are color coded accordingly in Figure~\ref{fig:invariance-summary}.
 \vspace{-1em}
\begin{center}
\begin{minipage}{1\linewidth}
\centering
\includegraphics[width=\linewidth]{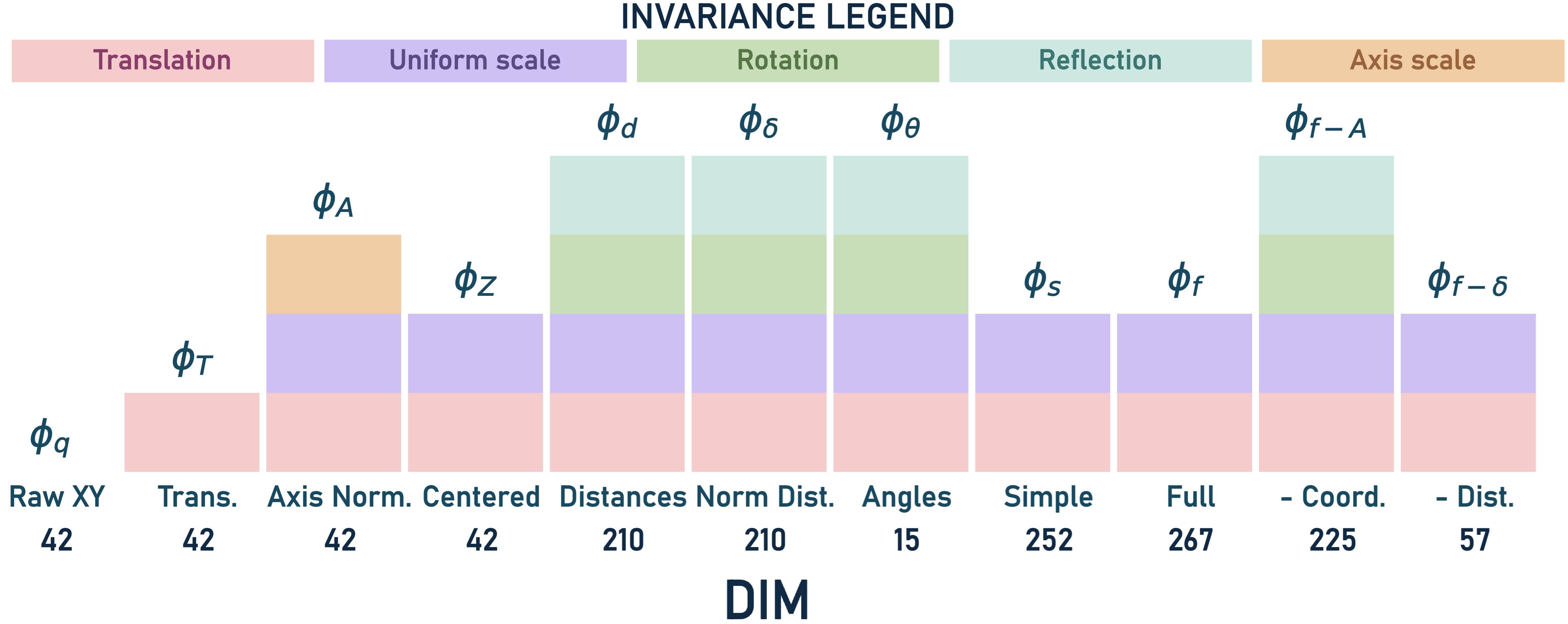}
\vspace{-2em}
\captionof{figure}{Visual Summary of Invariances.}
\label{fig:invariance-summary}
\end{minipage}
\vspace{-1.2em}
\end{center}
}
These feature spaces were curated around one simple principle: remove variation that is irrelevant to the classification task whilst retaining class discriminative geometry. This follows the broader computer vision principle of constructing descriptors around the reduction of sensitivity to irrelevant transformations \citep{Lowe2004}. Whilst the mapping from $\mathbb{R}^{H\times W\times C}$ to $\mathbb{R}^{21\times d}$, achieved through MediaPipe Hands, removes a substantial amount of visual nuisance, we do retain variation associated with scale and translation, both irrelevant to the classification task. Put differently, the size and location of the hand are not informative for identifying a gesture. This becomes evident when we observe pragmatic data, collected under conditions that mirror the typical deployment environments these recognition systems are expected to operate in.
\vspace{-1.9em}
\begin{center}
\begin{minipage}{0.95\linewidth}
\centering
\includegraphics[width=\linewidth]{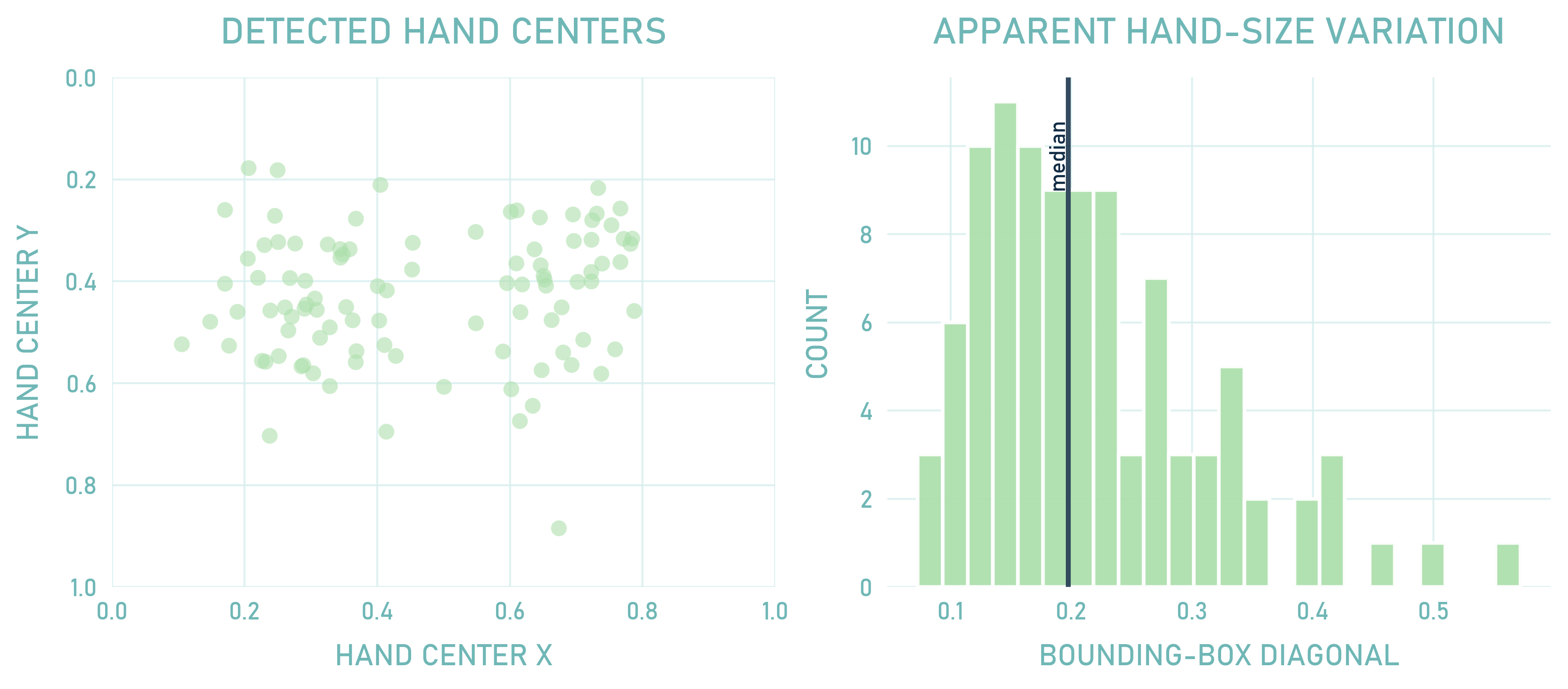}
\vspace{-2em}
 \captionof{figure}{Realistic Variation.}
 \label{fig:hagrid-position-scale} 
 \end{minipage}
 \vspace{-0.8em}
 \end{center}
 Figure~\ref{fig:hagrid-position-scale} illustrates this issue for a representative HaGRID \citep{Kapitanov2024HaGRID} gesture class. With the left figure being the detected hand centers, and the right being apparent hand-size variation. The hand appears in different regions of the frame and at different apparent sizes, even within a single isolated gesture class, implying that the variation shown here is not class discriminative. Therefore, within this space, any classifier $g$ is asked to solve two problems at once: identify the gesture and account for the location and size of the hand. From a learning efficiency perspective, this is inefficient. $g$ should, ideally,  be encouraged to exclusively model gesture discriminative geometry rather than nuisance variation in image location or object scale.
 \grev{This study employs image-corrected landmark configurations that form the common basis for all representations and must satisfy shared usability criteria: $\min\{R_x,R_y,D_i,\min_{(a,b)\in\mathcal E_\theta}\|q_a-q_b\|_2
\}>10^{-6}$ $\&$ $
Q,\ \{\phi_r(Q)\}_r\ \text{finite}$}
\grev{, where $\mathcal E_\theta$ denotes the set of segment pairs used in the angular construction. Any violation renders the sample unusable across all representations, regardless of individual computability. Such samples remain in the evaluation denominator, without attempts to repair degenerate configurations.
Translation and positive uniform scale are treated as nuisance transformations for the selected static-gesture task \citep{Lowe2004}. Under $q_j'=cq_j+t$, $c>0$, we have $b'=cb+t$, $\Lambda'=c\Lambda$ and $D'=cD$. Therefore $\phi_A,\phi_\delta,\phi_\theta,\phi_Z,\phi_s$ and $\phi_f$ are invariant wherever both evaluations remain usable. The translated map $\phi_T$ retains size, and unnormalized distances scale by $c$. A formal justification of these claims are included in the supplement material \citep{Mauree2026MathematicalSupplement}}
Whilst it might be tempting to indiscriminately remove all affine variation, not every transformation is irrelevant to the classification task. In our setting, translation and scale are treated as primary nuisance effects, whereas rotation is deliberately preserved since hand orientation can itself be gesture-discriminative in sign language. Although $\phi_d$ and $\phi_\theta$ provide rotation-invariant internal geometry, complete affine invariance is not explicitly imposed. The key implication is that different feature maps $\phi_r$ encode specific\grev{, not necessarily independent, }assumptions regarding the relevance of any particular transformation to the classification task. The goal is therefore not simply to reduce the number of variables, but to construct feature spaces that preserve gesture-relevant geometry whilst reducing sensitivity to irrelevant effects.

\vspace{-0.5em}
\section{Study Design and Evaluation Protocol}
\label{sec:experiment protocol}

\paragraph{\grev{Data And Model Fitting:}}
\grev{The protocol is executed on the HaGRID subset discussed in section \ref{sec:landmark feature representations}. Landmarks are extracted using MediaPipe $0.10.21$ in static-image mode, allowing up to two detected hands per image. Where multiple hands are detected, annotation-assisted matching identifies the target hand using bounding-box overlap, requiring an intersection-over-union  of at least $0.10$ and a margin greater than $0.05$ over the next candidate. Rather than being discarded, images with detection, matching, or feature extraction failures are retained as unsuccessful cases so as to allow for a performance metric reflective of both reliability of landmark extraction and the subsequent classification. Evaluation uses fixed, participant-disjoint partitions comprising $2,000$ TRAIN participants ($11,339$ images), $500$ VAL participants ($4,467$ images), and $1,000$ TEST participants ($15,298$ images). TRAIN is reserved for preprocessing, hyperparameter selection, and model fitting, whilst VAL is used for representation comparison and developmental decision making. TEST is used to assess the generalization to unseen participants. Models and development choices are fixed before TEST evaluation, without refitting on TRAIN+VAL.} We use two standard classifiers: \grev{multinomial logistic regression (LR) and random forest (RF)}. LR serves as a linear baseline. Strong performance under this model would suggest that the representations facilitate approximately linear separation of gesture classes. Predictors are standardized before fitting as features inherently have differing scales. We use RF to provide a foil model capable of capturing non-linear interactions amongst features without explicit specification \citep{Biau2016}. The choice of classifiers is fairly discretionary as these were meant not to exhaust the space of possible models, but to provide a controlled setting in which differences can be attributed primarily to the feature representation. \grev{Hyperparameters are selected using three-fold participant-grouped cross-validation on TRAIN, with a common seed. For LR, the regularization parameter is chosen from $\text{C}\in\{0.01,0.1,1,10,100\}$. RF employs 300 trees and searches twelve configurations varying maximum depth, minimum leaf size, and feature subsampling. Primary accuracy is computed by averaging correctness over all selected images within each participant, counting failed observations as incorrect, before averaging equally across participants. Hyperparameter selection maximizes participant-balanced out-of-fold accuracy, accounting for failed observations, whilst final models are fitted using usable TRAIN images only.}
\vspace{-0.8em}
\paragraph{\grev{Representation Selection and Evaluation: }}
\grev{Representation evaluation is performed separately for each classifier over the fixed nine-map pool in Figure \ref{fig:invariance-summary}. In line with design philosophy discussed in Section~\ref{sec:introduction}, the  following analysis aims to examine the three complementary aspect central to the problem: clean classification performance, robustness under image degradation, and the predictive contribution of individual feature components. Whilst the first two are assessed through the present evaluation, the latter is investigated through the ablation experiments in Section~\ref{sec:hybrid}, where we motivate and further distinguish the availability of geometric information from predictive utility. We first assess clean classification accuracy using the original, unmodified images. This evaluation metric defines the predictive performance of each representation under the original image conditions. This notion of CLEAN accuracy is assessed on the full TEST cohort, with detection matching, and feature extraction failures retained as unsuccessful cases. The resulting metric is a performance measure of the complete classification pipeline inclusive of landmark extraction reliability rather than being conditional to successful extraction.Whilst a strong showing under the CLEAN metric might infer that a representation is predictively suggestive, it does not necessarily warrant that a similar performance extrapolation will persist under more pragmatic image conditions. In particular, geometric invariance does not inherently or intuitively guarantee robustness to landmark location perturbations derived from pre-detection image corruption}To unambiguously \grev{investigate} whether the invariance properties translate to empirical robustness, \grev{we introduce four synthetic image corruptions before landmark detection, following the broader framework of corruption-based robustness evaluation \citep{HendrycksDietterich2019}.

\begin{center}
\vspace{-1.5em}
\begin{minipage}{1.02\linewidth}
\centering
\includegraphics[width=0.9\linewidth]{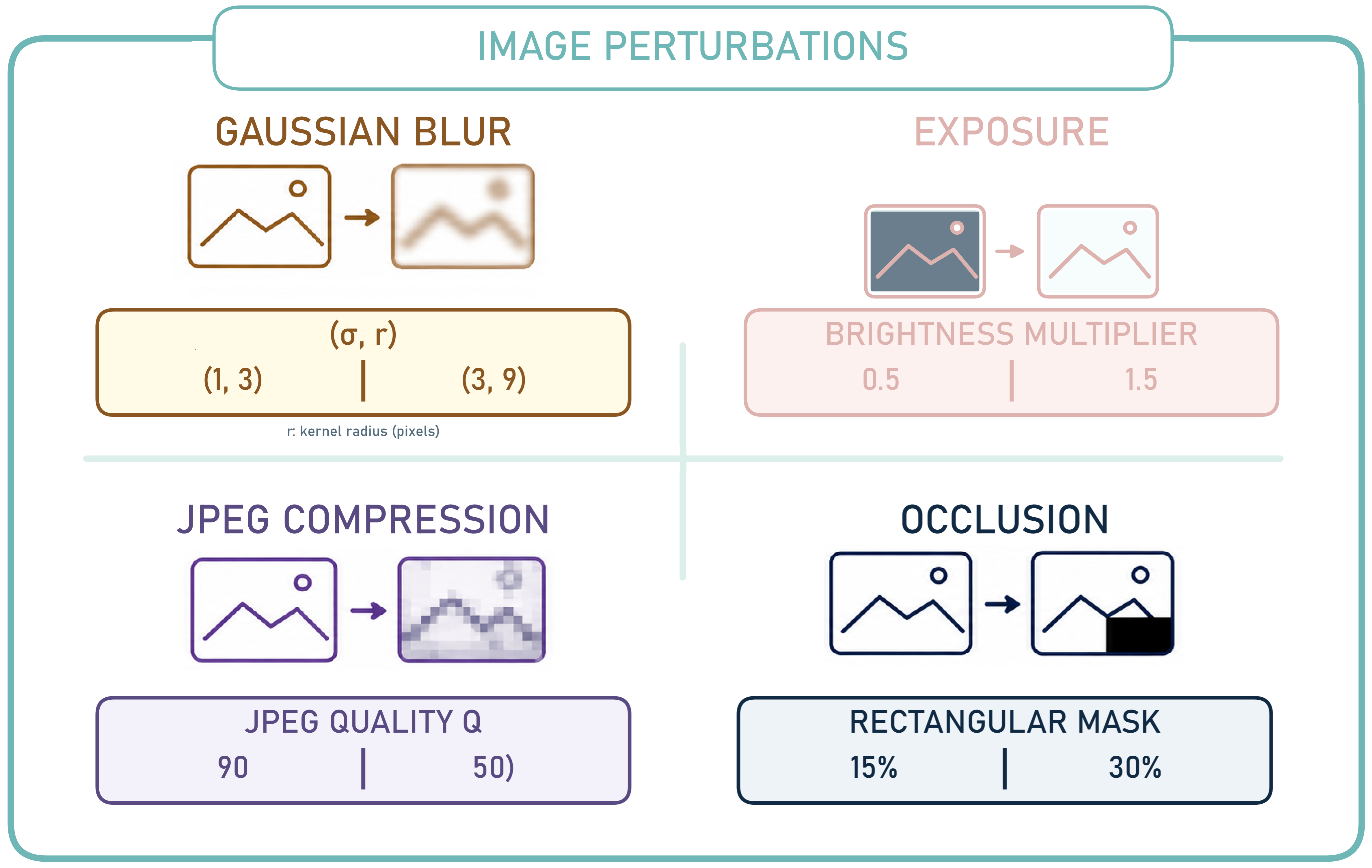}
 \vspace{-1.0em}
 \captionof{figure}{Perturbation Families.}
 \label{fig:Perturbation Families} 
 \end{minipage}
 \end{center}
  \vspace{-1em}
 Gaussian blur where $r$ denotes kernel radius; exposure changes with brightness multipliers; JPEG compression at quality level; and black rectangular occlusions covering a percentage of both image width and height, respectively. Each family is evaluated at two severity settings denoted in Figure~\ref{fig:Perturbation Families}. For occlusion, three fixed random placements are generated per severity level and shared across representations to ensure paired comparisons. Blur, exposure, and JPEG compression are deterministic at each severity level. These corruption settings are not treated as independent statistical replications. Robustness is appraised on a fixed subset of $500$ VAL and $500$ TEST participants, retaining at most $5$ images per participant for a total of $2242$ images.To ensure uniform comparability across perturbation families and severity settings, all corruption configurations are assessed on the same fixed set of images within the respective VAL and TEST cohorts. Their unmodified counterparts serve as the reference for measuring corruption-induced degradation. Beyond a considerable reduction in computational cost, this fixed assessment cohort precludes the attribution of observed performance differences to variations in the selected images or participants. Importantly, these corruptions are introduced upstream pre-detection and exclusively during evaluation, foregoing any further model fitting. Consequently, the observed performance not merely measures the sensitivity of the geometric features themselves, but the combined effects of realistic image degradation, landmark extraction reliability, feature construction, and classification. In this sense, being a true qualification of  end-to-end robustness.}
\vspace{-0.8em}
\paragraph{\grev{Statistical analysis: }}
\grev{To quantify the predictive performance of each representation, we consider three complementary measures of accuracy: participant-balanced classification accuracy, paired differences between representations, and clean-to-corrupted accuracy degradation. All the following measurements are computed at the participant level, avoiding a disproportionate participant weight relative to their image contribution count.Let $m$, $r$, $c$, and $p$ denote the classifier, representation, evaluation condition, and participant, respectively. For participant $p$, accuracy is defined as
\begin{center}
\vspace{-0.6em}
\refstepcounter{equation}
\resizebox{0.6\linewidth}{!}{$
a_{m,r,c,p}
=
\frac{1}{n_p}
\sum_{i\in\mathcal I_p}
\mathbf{1}
\{\widehat y_{m,r,c,i}=y_i\},
\qquad (\theequation)
$}
\vspace{-0.6em}
\end{center}
where $\mathcal I_p$ contains the $n_p$ selected images belonging to participant $p$, and unsuccessful pipeline observations contribute zero. The corresponding participant-balanced accuracy is
\begin{center}
\vspace{-0.5em}
\refstepcounter{equation}
\resizebox{0.5\linewidth}{!}{$
A_{m,r,c}
=
\frac{1}{N}
\sum_{p=1}^{N}a_{m,r,c,p},
\qquad (\theequation)
$}
\vspace{-0.5em}
\end{center}
where $N$ denotes the number of evaluated participants. For clean evaluation, $c=0$ represents unmodified images, whilst corruption conditions specify the perturbation family, severity, and realization. Family-level accuracy is obtained by averaging across the prescribed settings and realizations within each participant before aggregation across participants.To compare representations $r$ and $\tilde{r}$ evaluated on the same participants, we define the paired accuracy difference
\begin{center}
\vspace{-0.5em}
\refstepcounter{equation}
\resizebox{0.6\linewidth}{!}{$
\Delta_{m,r,\tilde{r},c}
=
\frac{1}{N}
\sum_{p=1}^{N}
\left(
a_{m,r,c,p}-a_{m,\tilde{r},c,p}
\right).
\qquad (\theequation)
$}
\vspace{-0.5em}
\end{center}
Here positive values favour representation $r$, whilst negative values favour $\tilde{r}$. Corruption-induced degradation is quantified through
\begin{center}
\vspace{-1.5em}
\refstepcounter{equation}
\resizebox{0.4\linewidth}{!}{$
D_{m,r,c}
=
A_{m,r,0}-A_{m,r,c},
\qquad (\theequation)
$}
\vspace{-0.8em}
\end{center}
where both accuracies are evaluated on the same fixed subset of participants and images. Positive values indicate a reduction in performance relative to the corresponding clean baseline.\grev{ The statistical uncertainty of these measures is quantified using $10,000$ participant-block bootstrap following the principles of clustered-data
resampling \citep{FieldWelsh2007}.. For each resample, $N$ participants are randomly sampled with replacement, retaining their complete observations across representations and evaluation configurations. Participant-balanced accuracy is then recomputed by averaging the sampled participant-level accuracies, with repeated participants contributing upon each reselection. The randomness introduced through this resampling produces an empirical distribution of $10,000$ accuracy estimates, whose $2.5$th and $97.5$th percentiles define the individual, unadjusted $95\%$ confidence intervals. The same participant blocks are resampled jointly when evaluating paired representation differences or clean-to-corrupted degradation, preserving the pairing of observations whilst accounting for within-participant dependence. The resulting intervals quantify sampling uncertainty conditional on the fitted models and selected evaluation protocol, without incorporating variability arising from retraining or hyperparameter selection.}

Supplementary post hoc hypothesis tests are additionally considered to assess systematic differences amongst representations. Whilst the participant-block bootstrap already provides uncertainty estimates for paired accuracy differences, rank-based tests offer a computationally less demanding alternative when extensive resampling is impractical. For each classifier, Friedman tests compare all nine representations on clean TEST ($N=1,000$) and separately within each corruption family ($N=500$), using participant-level ranks \citep{Friedman1937}. Two-sided Wilcoxon signed-rank tests subsequently assess individual paired differences \citep{Wilcoxon1945}, with a single Holm correction controlling the family-wise error rate across all hypotheses \citep{Holm1979}. These tests are supplementary to the bootstrap analysis and do not directly quantify uncertainty in mean accuracy differences. Their full results and implementation details are provided in Supplement~\citep{Mauree2026StatisticalSupplement}.}
\vspace{-1.2em}

\section{Results and Discussion}
\label{sec:results}
The results reveal an apparent inference: geometric feature selection has a substantial influence on classification performance. 
Raw coordinates perform substantially worse. In contrast, representations that suppress translation or scale variances, or encode internal hand geometry, \grev{achieve stronger predictive performance}. 
Figure~\ref{fig:hagrid_accuracy} isolates this effect through a \grev{plot of the participant-balanced accuracy across nine geometric representations with the shaded regions indicating the individual $95\%$ participant-block bootstrap
confidence intervals.}
 \vspace{-1.8em}
\begin{center}
\hspace*{-0.03\linewidth}
\begin{minipage}{1.1\linewidth}
\centering
\includegraphics[width=\linewidth]{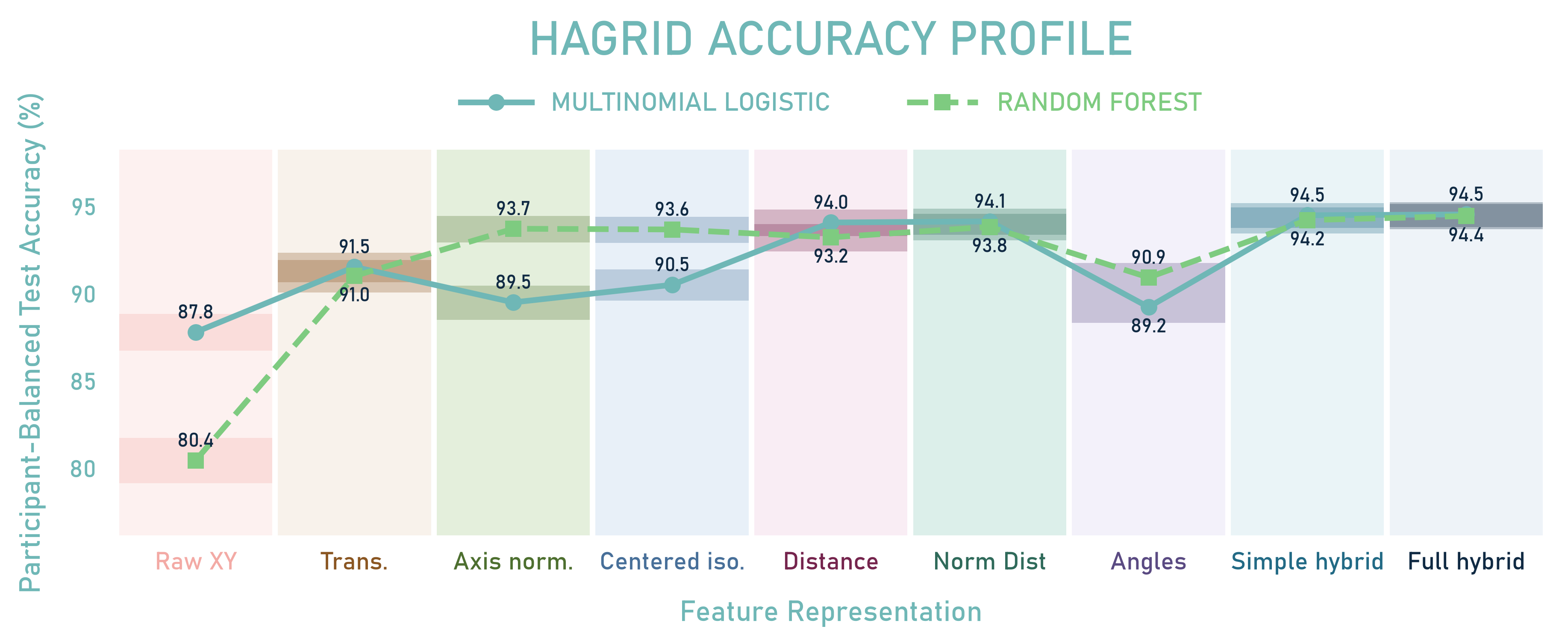}
\vspace{-2em}
\captionof{figure}{CLEAN Accuracy Profile.}
\label{fig:hagrid_accuracy}
\end{minipage}
\vspace{-0.8em}
\end{center}
\grev{Through the evaluation, where landmark detection ($3.14 \%)$ and target matching ($1.26\%$) were retained in the accuracy denominator as unsuccessful cases,} an observable accuracy hierarchy 
suggests some form of correlation between geometric abstraction and classification performance. As geometrically richer information is encoded, the gesture classes become more separable and \grev{predictive performance can improve. However, the limited differences between the two hybrid representations indicate that additional features do not necessarily yield substantial predictive gains. } \grev{$\phi_\theta$} provides an instructive foil. Although angles are invariant to translation and scale, they perform worse than other geometric alternatives, implying that local bending and orientation alone are not discriminative enough for classification. In gaining invariance, they discard spacing and broader hand shape information that are integral to certain gestures.\grev{  The HaGRID results suggest that geometric representations may offer greater robustness under realistic image condition variation. However, because several sources of variation occur simultaneously these observations alone cannot establish which factors account for the observed differences in performance}.  Thus, we \grev{consider the} synthetic perturbation experiment.
\vspace{-1.6em}
\begin{center}
\hspace*{-0.07\linewidth}
\begin{minipage}{1.1\linewidth}
\centering
\includegraphics[width=\linewidth]{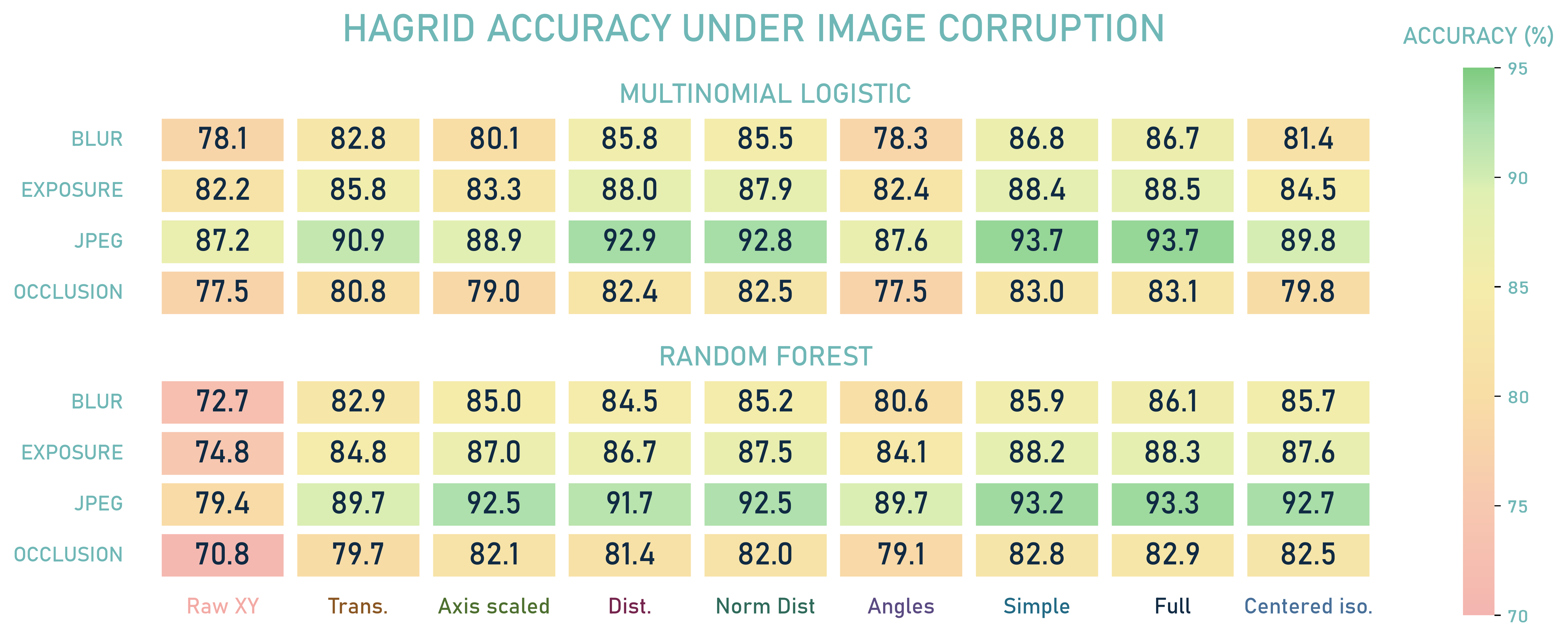}
\vspace{-2em}
\captionof{figure}{Corruption Heatmap}
\label{fig:robustness_experiment}
\end{minipage}
\vspace{-0.9em}
\end{center}
This experiment is not merely an additional accuracy test; it instead probes the task-dependent nature of feature construction by asking not which invariances can be imposed but rather which sources of variation are irrelevant to the problem. \grev{Figure~\ref{fig:robustness_experiment} presents the participant-balanced TEST accuracy, as defined in Section~\ref{sec:experiment protocol}, averaged across the two severity settings and three prescribed draws for each corruption family, separately for LR and RF. Each reported accuracy therefore reflects the mean performance across participants, rather than individual images, with detection and matching failures retained as unsuccessful cases. Across the $24$ corruption configurations, image-level landmark detection failure rates ranged from $3.39\%$ to $14.32\%$, whilst target matching failures ranged from $0.67\%$ to $5.35\%$.} Figure~\ref{fig:robustness_experiment} shows a clear separation between raw coordinates and invariant features. $\phi_q$ performs substantially worse, achieving only $78.1\%$ and $72.7\%$ under Gaussian blur for LR and RF, respectively. In contrast, $\phi_d,\phi_\delta,\phi_s,\phi_f$ remain amongst the strongest performers across all perturbation types, with $\phi_f$ achieving $86.7\%$ and $86.1\%$ under the same conditions, whilst maintaining strong performance across the remaining corruption families. \grev{Relative to $\phi_q$, this corresponds to an observed advantage of approximately $8.6$ and $13.4$ PP for LR and RF, respectively. Whilst these differences illustrate the predictive advantage of the full hybrid under Gaussian blur, their uncertainty must be considered through paired participant-block comparisons rather than the individual accuracy estimates alone. The paired bootstrap comparison between $\phi_f$ and $\phi_s$ reveals considerably smaller differences, with $\phi_f$ exhibiting a $0.107$ PP deficit for LR ($95\%$ CI: $[-0.247,0.040]$) and a $0.223$ PP advantage for RF ($95\%$ CI: $[0.033,0.440]$). Whilst the LR difference of $-0.107$ PP is not significant, the $0.223$ percentage-point advantage for RF is significant under the unadjusted confidence level, however, this difference does not remain statistically significant after Holm correction in the supplementary hypothesis tests. These observations suggest that the predictive advantages of geometric feature construction persist under image degradation, although additional geometric descriptors do not necessarily translate to substantial improvements in robustness.}
\begin{strip}
\vspace{-3em}
\begin{center}
\includegraphics[width=0.93\textwidth]{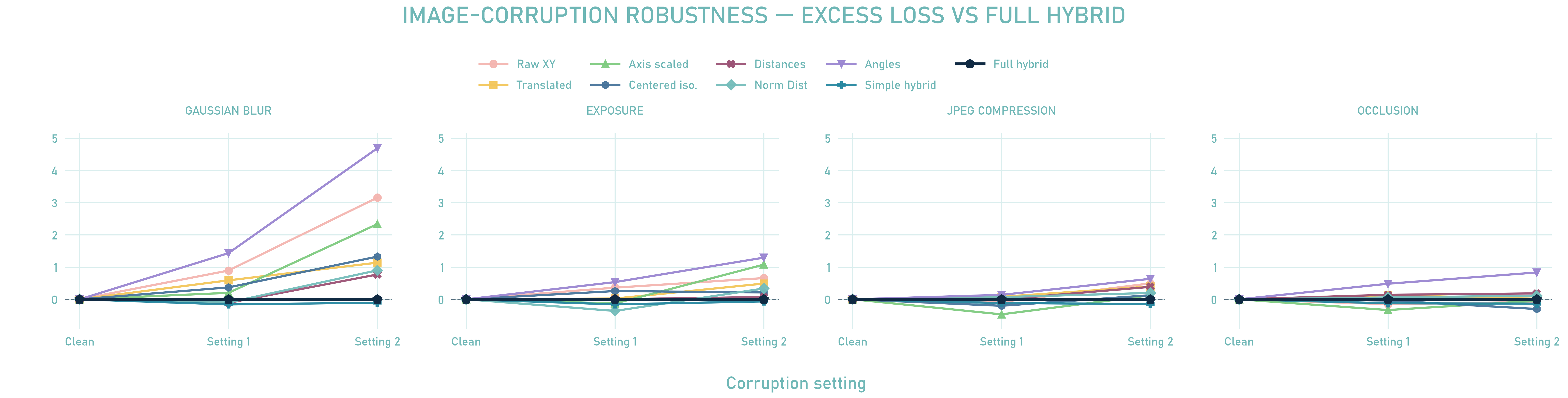}
\vspace{-1em}
\captionsetup{type=figure}
\captionof{figure}{Feature Visualisation-Summary And Justification of feature families considered}
\label{fig:corruption excess loss}
\vspace{-1.6em}
\end{center}
\vspace{-0.8em}
\end{strip}
\grev{Whilst Figure~\ref{fig:robustness_experiment} establishes the classification performance of each representation under image degradation, it does not distinguish differences in baseline predictive performance from the relative degradation induced by each perturbation. To investigate this distinction, Figure~\ref{fig:corruption excess loss} examines the excess relative accuracy loss with respect to the full hybrid $\phi_f$. Extending the absolute accuracy loss defined in Section~\ref{sec:experiment protocol}, let $A_{r,m}^{(0)}$ denote the clean accuracy of representation $r$ under classifier $m$ on the matched robustness subset, and let $A_{r,m,c,s}$ denote its accuracy under corruption family $c$ at severity setting $s$, averaged across the three prescribed draws. We define the relative percentage accuracy loss as
\begin{center}
\vspace{-1.6em}
\refstepcounter{equation}
\resizebox{0.6\linewidth}{!}{$
L_{r,m,c,s}
=100\left(A_{r,m}^{(0)}-A_{r,m,c,s}\right){A_{r,m}^{(0)}}^{-1}.
\qquad (\theequation)
$}
\vspace{-0.5em}
\end{center}
Averaging equally across (LR) and (RF), we obtain
\begin{center}
\vspace{-0.5em}
\refstepcounter{equation}
\resizebox{0.5\linewidth}{!}{$
\overline L_{r,c,s}
=\left(L_{r,\mathrm{LR},c,s}+L_{r,\mathrm{RF},c,s}\right)2^{-1}.
\qquad (\theequation)
$}
\vspace{-0.5em}
\end{center}
The excess loss relative to the full hybrid is then defined as
\begin{center}
\vspace{-0.7em}
\refstepcounter{equation}
\resizebox{0.4\linewidth}{!}{$
E_{r,c,s}
=\overline L_{r,c,s}-\overline L_{\phi_f,c,s}
\qquad (\theequation)
$}
\vspace{-0.6em}
\end{center}
Thus, $\phi_f$ serves as the zero reference, with positive values indicating greater relative accuracy degradation than $\phi_f$ and negative values indicating smaller degradation. These differences are expressed in percentage points (PP) of relative accuracy loss, rather than classification accuracy. Under Gaussian blur, the angular representation $\phi_\theta$ exhibits the greatest excess loss, reaching $4.69$ at the second setting, compared with $3.16$ for $\phi_q$ and $2.33$ for $\phi_A$. In contrast, $\phi_d$ and $\phi_\delta$ exhibit smaller excess losses of $0.77$ and $0.90$, respectively. The remaining perturbation families reveal a somewhat differing pattern. Under exposure changes, $\phi_A$ and $\phi_\theta$ exhibit a family-averaged excess loss of $0.91$ and $0.79$ PP respectively, whilst the distance-based and hybrid representations remain comparatively close to $\phi_f$.} 
\grev{At the evaluated compression levels, it is clear that the hand's spatial structure remains sufficiently preserved thus limiting the perceivable differences across representations.} \grev{On the flip side of the coin, occlusion enforces the lowers absolute classification accuracies, yet somewhat counterintuitively produces relatively small excess loss with $\phi_\theta$ exhibiting the largest family-averaged excess loss of $0.54$ PP.} \grev{Unlike the other families, occlusion directly conceals portions of the hand which inherently could introduce errors at the extraction stage that subsequently propagate across all geometric representations. Thus this loss in accuracy from retaining unsuccessful extractions are paradoxically paired with modest differences in excess loss.} \grev{Perhaps more intriguingly, across all four perturbation types $\phi_s$ remains remarkably close to $\phi_f$, with slightly smaller family-averaged relative losses.}
\grev{Across the perturbation families, the distance-based and hybrid representations generally exhibit comparatively modest excess losses. From a geometric perspective, this suggests that preserving relative landmark spacing, particularly when complemented by normalized spatial coordinates, may offer predictive advantages under image degradation. However, the extent of such robustness cannot be inferred from geometric invariance alone, as representations sharing similar invariance properties need not infer comparable sensitivity to perturbations. In particular} $\phi_\theta$ exemplifies this inference. \grev{Despite sharing translation and scale invariance with $\phi_\delta$, it exhibits substantially greater excess loss under Gaussian blur.} This observation is consistent with the sensitivity of angular features to landmark-localization errors. From the definition of $\theta_{abc}$ in Eq.~\ref{eq:angle}, small changes in landmark positions can alter the directions of the vectors used to compute joint angles, particularly for nearly aligned segments. \grev{Nevertheless, since the experiment assesses the complete classification pipeline, the observed degradation cannot be attributed exclusively to this mechanism.} Thus, invariance to translation and scale does not inherently guarantee robustness to image degradation or errors introduced during landmark detection. \grev{Supplementary post hoc Friedman and Wilcoxon tests further support differences amongst representations, with Holm-adjusted significance observed across multiple comparisons. However, the full and simple hybrids bare no statistically significant differences after adjustment. Complete statistical results are provided in Supplement~\citep{Mauree2026StatisticalSupplement}}

\section{Building A Hybrid Representation}
\label{sec:hybrid}
A fair criticism, thus far,  would be that the hybrid representation may appear to have been chosen heuristically. Whilst the choice of features was initially motivated by the complementary landmark geometry they preserved, this intuition becomes less dependable for problems where the relevant structure is not visually obvious. \grev{Interestingly, despite $\phi_f$ explicitly incorporating angular features, $\phi_s$ is comparable. Which raises a natural question: does the addition of features encode genuinely new geometry, or do they merely re-express existing information? To investigate this distinction, we first characterize the geometry preserved through the simple hybrid acknowledging that from an information encoding standpoint whilst each component might describe different aspects of the same landmark configuration, they need not contain independent geometric information.
This motivates the following proposition, establishing the relationship between the simple hybrid $\phi_s$ and the centered, isotropically normalized representation $\phi_Z$. 
\vspace{-0.3em}
\begin{proposition}[Equivalent geometric encodings]
\label{prop:geometric_equivalence}
For noncollinear, usable planar landmark configurations,
the simple hybrid representation
$\phi_s=(\phi_A,\phi_\delta)$ and the centered,
isotropically normalized representation $\phi_Z$
determine one another. Consequently, both identify
the landmark configuration up to translation and
positive uniform scaling, without identifying
rotations or reflections.
\end{proposition}
\vspace{-0.3em}
A more formal definition and proof of this proposition is provided in the Supplement~\citep{Mauree2026MathematicalSupplement}. Consequently, despite their differing ambient dimensionalities of $252$ and $42$, respectively, both representations preserve equivalent geometric information on its noncollinear, realizable domain, within the broader translation-and-scale quotient spaced numerically realized through $\phi_Z$. 
Furthermore, the angular features in $\phi_f$ are recoverable from normalized pairwise distances \citep{Mauree2026MathematicalSupplement}. Hence, appending the angular block introduces no additional geometric information. So, if both representations encode the same geometric information, why does the $252$-dimensional simple hybrid outperform its more succinct counterpart? Simply put, whilst two representations may preserve the same information, the extent to which a classifier can exploit is highly situational. For instance, reconstructing angular features from normalized distances requires nonlinear operations, which are not automatically available to a linear classifier. Explicitly supplying these features may therefore improve predictive accessibility without introducing new geometric information. This distinction between information preservation and predictive utility motivates the subsequent ablation experiments, through which we investigate the contribution of each constituent component to classification performance.
}\vspace{-0.4em}
\begin{center}
\begin{minipage}{0.99\linewidth}
\centering
\includegraphics[width=\linewidth]{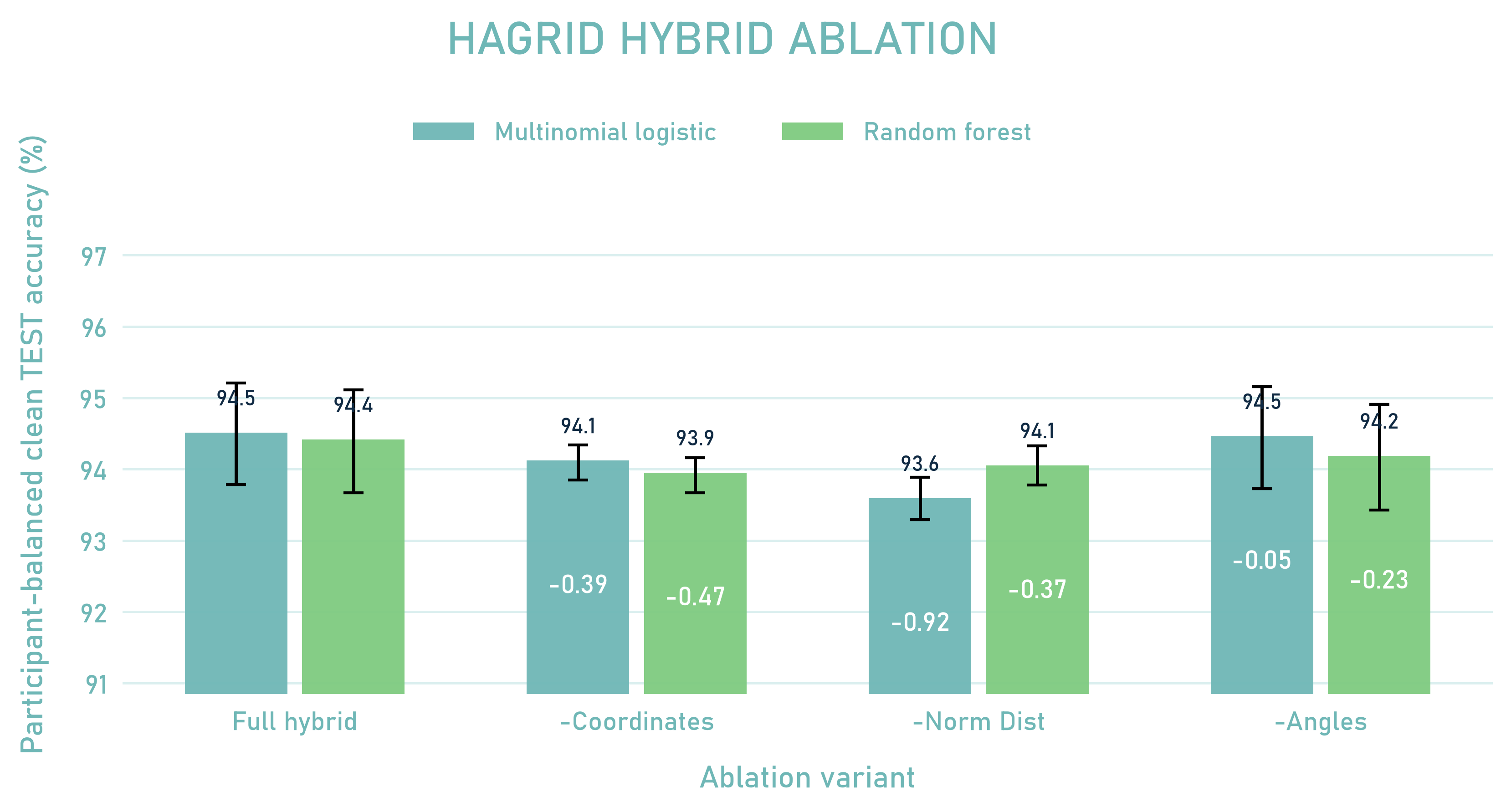}
\vspace{-2em}
\captionof{figure}{Accuracy Across Ablations.}
\label{fig:ablation_experiment}
\end{minipage}
\vspace{-0.5em}
\end{center} To do so, we compare the full, intuitively built, hybrid representation \grev{$\phi_f$} against three ablated variants formed by removing \grev{axis-normalized} coordinates, normalized distances, or angle features, using the same \grev{participant-disjoint TEST protocol} employed for the main HaGRID experiment. For each ablation, the change in performance relative to \grev{$\phi_f$} estimates the contribution of the removed component under the same dataset and  protocol. A substantial drop indicates that the omitted feature \grev{improves predictive performance under the chosen classifier and fitting procedure, even when its information is recoverable from other components}; a negligible change suggests some form of redundancy or perhaps limited marginal value. Figure~\ref{fig:ablation_experiment} shows that \grev{the full hybrid $\phi_f$ achieves the highest clean TEST accuracy under both classifiers, although its advantage over the simple hybrid $\phi_s$ is small.} The largest performance drops, across either model, occur when \grev{$\phi_A$ or $\phi_\delta$} are excised, indicating that \grev{$\phi_f$'s} predictive aptitude can be largely attributed to these features. For \grev{LR}, removing \grev{$\phi_\delta$ reduces accuracy by $0.919$ PP, whilst removing $\phi_A$ results in a $0.390$ PP decrease}. For RF, the corresponding reductions are \grev{$0.366$ and $0.467$ PP}, respectively. This similarity in behavior across two distinct families of classifiers suggests that \grev{both components contribute to predictive performance, although their relative importance depends on the classifier.}
Removing \grev{$\phi_\theta$ produces smaller losses of $0.050$ and $0.231$ PP for LR and RF, respectively. Whilst angular features provide an interpretable description of local finger articulation, their limited marginal contribution is consistent with their geometric recoverability from the normalized-distance block.} Unsurprisingly, these findings are congruent \grev{with the broader patterns observed in} the earlier perturbation experiments. \grev{The ablation analysis suggests that much of the hybrid's predictive advantage arises from combining normalized hand layout and relative landmark spacing, whilst the angular block offers comparatively limited additional benefit under the evaluated conditions.} We therefore have this complementary dynamic, where for each feature, perturbation identifies the response to controlled variation, whilst ablation determines whether \grev{explicitly supplying the component contributes to predictive performance}.

This naturally suggests a general procedure for the construction of hybrid representations in other structured classification problems. First, candidate features can be derived from the available semantic structures like contours, segmentations, and anatomical regions. Whilst the initial set may be deliberately broad, inclusion in the final representation should be guided by the role each feature plays in the task. In some settings, rotation or shear may be nuisance variation, whilst in others, such as sign language, orientation may carry class discriminative information. Once these roles are identified, components with distinct geometric purposes can be combined into a provisional hybrid representation. Afterwards, perturbation and ablation experiments can be used to evaluate the composition. Where relevant geometry is less obvious, this procedure may be implemented incrementally. Candidate features can be added or removed one at a time according to \grev{validation performance}, robustness under perturbation, and the interpretability of the retained features. Hybrid construction thus becomes an empirical design procedure rather than an exercise in intuitive geometric perception.

\section{Illustrative Transfer to Anatomical Landmarks}
\begin{strip}
\vspace{-3.2 em}
\begin{center}
\includegraphics[width=0.93\textwidth]{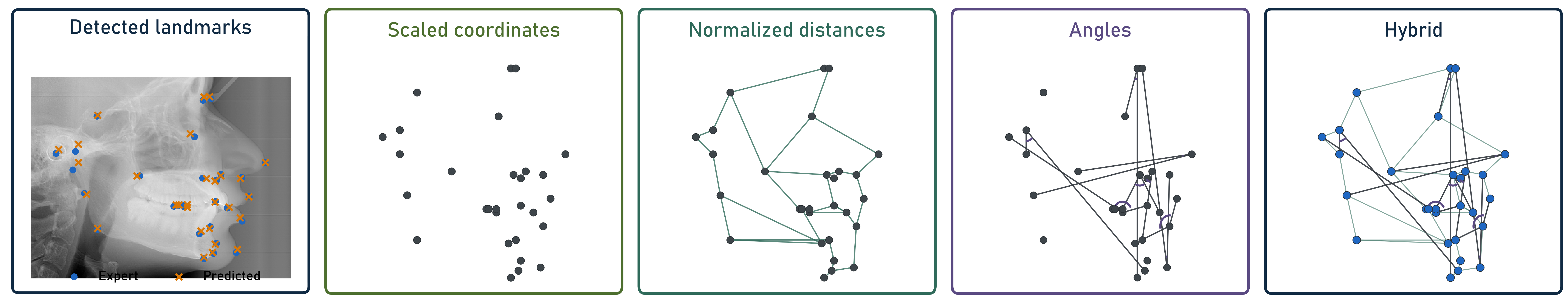}
\vspace{-0.4em}
\captionsetup{type=figure}
\captionof{figure}{Cephalometric Feature Representation Spaces}
\label{fig:Cephalometric Extension}
\vspace{-2.3em}
\end{center}
\end{strip}
\grev{To investigate the portability} of the proposed framework beyond hand landmarks or MediaPipe extraction, we considered a lateral cephalometric radiograph dataset \citep{Khalid2025AarizCephalometric} We implemented and trained a lightweight convolutional landmark detector \grev{to extract} $29$ anatomical landmarks, yielding the desired $L(I)=\{p_j\}_{j=1}^{29}$, $p_j\in\mathbb{R}^2$ mathematical structure. Thus the geometric constructions developed remains directly applicable despite the change in image domain and extraction mechanism. Figure~\ref{fig:Cephalometric Extension} illustrates the resulting pipeline on a radiograph from the official TEST partition. \grev{The detector achieved a mean radial error of $1.462$ mm ($95\%$ CI: $1.35$-$1.58$) across the $150$ TEST radiograph. Localization accuracy was further assessed through the successful detection metric: of the $4,350$ evaluated landmarks, $79.10\%$ were within $2$ mm and $95.26\%$ within $4$ mm. Beyond landmark localization, sufficiently derived anatomical measurements further demonstrated the interpretability of the extracted geometry, hence reinforcing the extension to geometric feature representations . Detailed localization results, and full reports of the measurement construction evaluation are provided in supplement~\citep{Mauree2026StatisticalSupplement}.} This extension illustrates the portability of the framework rather than establishing its predictive advantage in a second domain. 
\vspace{-0.5em}
\section{Conclusion}
This paper studied landmark-based feature construction as a problem-informed dimension reduction step for structured image classification. \grev{We argue that} the reduction of an image to its semantic landmark structure does not, \grev{by itself, resolve the problem of constructing effective representations for classification.} 
\grev{Through a participant-disjoint analysis of $9$ prototypical candidate maps, we were able to statistically demonstrate that variation in geometric encoding has a tangible influence on classification performance and response  end-to-end image degradation. More specifically, representations inclusive of normalized geometric structure generally inferred improved predictive efficacy, with varying magnitudes across perturbation types. This heterogeneity in performance was noticeably apparent under Gaussian blurring where discernible differences emerged between representations. On the other hand, occlusion captured a comparatively homogenueous performance despite inducing considerable end-to-end degradation. These findings delineate a fundamental distinction between theoretical geometric invariance and empirical robustness. Complementing these findings, is the geometric recoverability result that feature spaces of substantially differing dimensionalities can preserve equivalent information retrievable from their underlying basis configuration $Q$. Yet, this equivalence did not necessarily translate into comparable predictive performance, suggesting that the accessibility of geometric information is influenced by the chosen encoding and classifier. Ablations experiments further reinforced the predictive contributions of normalized spacial layout and relative landmark spacings. Whilst Limited differences between the simple and full hybrid representations suggest that additional geometric descriptors do not necessarily translate into appreciable predictive gains.}
The overarching implication is that dimensionality or accuracy alone are not sufficient criteria to comprehensively measure representation quality. \grev{A more comprehensive evaluation must consider the geometric information preserved, the variation suppressed, and the quantifiable accessibility made available to the classifier through the encoding.}Perturbation and ablation analyses consequently offer a complementary \grev{means evaluating} and validating \grev{the predictive utility of} post semantic structure extraction representations \grev{beyond clean classification accuracy.} 

Future work can extend this framework in two natural directions. First, the subsumption of temporal sequence modeling could extend beyond static encoding to incorporate motion. Second, whilst this study is \grev{restricted to realizable} explicit finite-dimensional Euclidean feature vectors a natural extension \grev{of the equivalence classes mentioned in Section~\ref{sec:hybrid}} over Hilbert spaces through kernel-based similarity learning \grev{could prove worthwhile}. As methods for semantic structure extraction become increasingly reliable across a more extensive range of classification problems, progress will gradually depend less on recovering the semantic structure and more on representing it effectively. The methodology proposed in this paper provides a systematic framework to address this nascent challenge.

\printcredits




\bibliographystyle{elsarticle-num}
\bibliography{cas-refs}

\end{document}